\documentclass[10pt, a4paper]{article}

\usepackage{lrec-coling2024} % this is the new style

\usepackage{multirow}
\usepackage{amsmath}
\usepackage{amssymb} % Add this line

\usepackage{booktabs} % for professional tables

\title{Improved Out-of-Scope Intent Classification with Dual Encoding and Threshold-based Re-Classification}

\name{Hossam M. Zawbaa\textsuperscript{\rm 1, \rm 2},
    Wael Rashwan\textsuperscript{\rm 1,  \rm 2, \rm 3},
    Sourav Dutta\textsuperscript{\rm 4}, 
    Haytham	Assem\textsuperscript{\rm 5}}
\address{
 \textsuperscript{\rm 1}Technological University Dublin, Ireland.\\
 \textsuperscript{\rm 2} Ireland’s Centre for Applied AI (CeADAR), Ireland\\
 \textsuperscript{\rm 3} 3S Group, Faculty of Business, TU DUBLIN, Ireland\\
    \textsuperscript{\rm 4}Huawei Research Centre, Ireland\\
    \textsuperscript{\rm 5}Amazon Alexa AI, United Kingdom\\
    hossam.zawbaa@gmail.com, wael.rashwan@tudublin.ie, sourav.dutta2@huawei.com,\\ hithsala@amazon.co.uk
}

\abstract{
Detecting out-of-scope user utterances is essential for task-oriented dialogues and intent classification. Current methodologies face difficulties with the unpredictable distribution of outliers and often rely on assumptions about data distributions.
We present the Dual Encoder for Threshold-Based Re-Classification (DETER) to address these challenges. This end-to-end framework efficiently detects out-of-scope intents without requiring assumptions on data distributions or additional post-processing steps. The core of DETER utilizes dual text encoders, the Universal Sentence Encoder (USE) and the Transformer-based Denoising AutoEncoder (TSDAE), to generate user utterance embeddings, which are classified through a branched neural architecture. Further, DETER generates synthetic outliers using self-supervision and incorporates out-of-scope phrases from open-domain datasets. This approach ensures a comprehensive training set for out-of-scope detection. Additionally, a threshold-based re-classification mechanism refines the model's initial predictions.
Evaluations on the CLINC-150, Stackoverflow, and Banking77 datasets demonstrate DETER's efficacy. Our model outperforms previous benchmarks, achieving an increase of up to 13\% and 5\% in F1 score for known and unknown intents on CLINC-150 and Stackoverflow, and 16\% for known and 24\% for unknown intents on Banking77. The source code has been released at \href{https://github.com/Hossam-Mohammed-tech/Intent_Classification_OOS}{https://github.com/Hossam-Mohammed-tech/Intent\_Classification\_OOS}.
\\ \newline }

\begin{document}

\maketitleabstract

\section{Introduction}
%Sourav-Added
%With the success of Generative AI models like ChatGPT, conversational frameworks are becoming the de-facto interactive interface for end-user assistance. Understanding the precise intents behind the user utterances constitutes a primary objective. A pivotal component of these systems is their ability to accurately interpret the underlying intent behind the user utterances. As such, intent detection becomes indispensable, guiding text-based chatbots in formulating appropriate responses \cite{Coucke_2018}. Given the vastness and unpredictability of possible user intents, it is challenging to have training data that encompasses the entire spectrum of queries and remarks.
%

The rise of Generative AI models such as ChatGPT underscores the prominence of conversational frameworks as the primary interactive interface for end-user assistance. A crucial goal within these systems is to precisely discern the user's intent behind their utterances, a fundamental component for formulating appropriate responses \cite{Coucke_2018}. However, the challenge lies in acquiring comprehensive training data that covers the diverse spectrum of potential user intents.

A competent intent classifier should identify the predefined intents and recognize {\em out-of-scope} (OOS) data. OOS data are inputs unrelated to the trained classes, representing queries that do not fit into predefined categories or intents \cite{Cavalin_2020}. This is essential for truly understanding a user's objective.

%% WR 
%While intent classifiers hold promise, their broader deployment is hindered by the need for domain-specific labelled datasets to train for rapid and widespread deployment. In essence, utilizing a successful intent classifier in low-data settings, typically observed in commercial systems, presents a significant challenge \cite{Casanueva_2020}.
%%WR
%In other words, \xcancel{allowing}\textcolor{red}{using} a successful intent classifier in low-data designs, typically in commercial systems, is a significant challenge \cite{Casanueva_2020}.
%%
%Modern Language models, like BERT, have been pivotal in this regard. They convert text (e.g., user utterances)  into vector representations, embedding rich semantic information that neural networks can leverage for efficient intent classification. Current strategies for OOS intent detection often lean on outlier detection, rooted in the known class decision boundaries or the model's classification confidence scores. However, these techniques are grounded in rigid assumptions on the input data distributions or require post-processing steps like tuning the confidence threshold. Further, such methods tend to exhibit limited generalisability on the OOS (unknown) intents due to limited training examples. To address this challenge, \citet{Zhan_2021} introduced synthetic outlier construction using in-domain (known) training data embeddings. However, these methods are static and do not offer opportunities for refining model predictions. 

Intent classifiers face challenges due to the requirement for domain-specific labelled datasets, particularly in low-data settings standard in commercial systems \cite{Casanueva_2020}. Modern Language models like BERT aid in efficient intent classification by converting text into vector representations rich in semantic information. However, current strategies for Out-Of-Scope (OOS) intent detection often lack generalizability and refinement opportunities \cite{Zhan_2021}. To address this, \citet{Zhan_2021} introduced synthetic outlier construction using in-domain training data embeddings, but these methods are static and offer limited potential for refining model predictions.

% Wael edited 
%In this paper, we introduce {\em Dual Encoder for Threshold-Based Re-Classification} (DETER), an efficient method for OOS intent detection without the need for data distribution assumptions or post-processing strategies. DETER synergizes two methods, confidence thresholding and synthetic outlier generation, offering enhanced intent classification performance.
% Wael edited 
%Our proposed deep neural network architecture for user intent classification uses a branched structure with dual sentence encoder representations. Specifically, we use the Universal Sentence Encoder (USE) ~\cite{Daniel_2018} and Transformer-based Sequential Denoising Auto-Encoder (TSDAE) ~\cite{Wang_2021}. As a preprocessing step, we fine-tune the TSDAE model in an unsupervised manner using only Clinc-150 in-domain text data without any labels. We then use these fixed representations and pass them through several dense layers with ReLU non-linear activations \cite{Maas_2013}. Finally, a softmax layer is used for multi-class classification. To enhance out-of-scope intent detection, our approach incorporates synthetic outlier generation. This technique amalgamates self-supervision with open-domain datasets, inspired by \citet{Zhan_2021}. 

In this paper, we introduce  {\em Dual Encoder for Threshold-Based Re-Classification} (DETER), a method for efficient Out-Of-Scope (OOS) intent detection. DETER combines confidence thresholding and synthetic outlier generation to enhance intent classification performance \cite{Zhan_2021, Daniel_2018, Wang_2021, Maas_2013}. Our neural network architecture utilizes dual sentence encoders: the Universal Sentence Encoder (USE) and Transformer-based Sequential Denoising Auto-Encoder (TSDAE) \cite{Daniel_2018, Wang_2021}. We fine-tune the TSDAE model unsupervisedly using Clinc-150 data and incorporate synthetic outlier generation to enhance OOS intent detection \cite{Zhan_2021}.

% Wael edited 
Summarizing our contributions, this research presents the innovative {\em Dual Encoder for Threshold-based Re-classification} (DETER) optimized for out-of-scope detection, offering a comprehensive solution devoid of stringent data distribution assumptions. The system adeptly captures an array of synaptic and semantic features by utilizing dual text encoding mechanisms, namely USE and TSDAE, thereby refining user utterance representations and amplifying the efficiency of out-of-scope detection. A distinctive threshold-based re-classification mechanism further augments prediction reliability, ensuring a consistent and precise identification of out-of-scope intents. In addition, DETER employs a nuanced approach to synthetic outlier generation, merging self-supervision with open-domain datasets to cultivate a diversified training set. Empirical assessments on CLINC-150, Stackoverflow, and Banking77 datasets underscore its superior performance, with notable improvements in F1 scores relative to established benchmarks. 
Furthermore, DETER's streamlined architecture, with only $1.5$ million trainable parameters, contrasts markedly with models like \citet{Zhan_2021}, boasting $125$ million parameters, showcasing enhanced computational efficiency and scalability without compromising performance.

\section{Related Literature}

%WR
The choice of embeddings significantly impacts intent classification, with sentence-level embeddings gaining precedence over token-level ones in Natural Language Understanding (NLU). However, many specialized NLU models struggle outside core domains. Graph-based representations, which translate nodes to vector embeddings, hold promise for enhancing intent detection, particularly in Out-Of-Scope (OOS) classification. The potential of synthetic datasets, both artificial and real-world derived, is being explored to improve model adaptability across diverse data distributions. Techniques such as temperature scaling and strategic perturbations aid in distinguishing known from unknown samples. While Transformer models like BERT and RoBERTa have revolutionized NLU with their OOS resilience, optimization remains crucial. Additionally, emerging pre-trained dual encoders like USE and ConveRT further augment intent detection, with evidence suggesting their superiority over refinements such as BERT-Large.

The selection of embeddings holds critical importance in intent classification. TEXTOIR, a recent platform introduced for intent detection, provides tools for open intent detection and discovery tasks, offering a practical toolbox with expandable interfaces \cite{zhang-etal-2021-textoir}. For this study, we utilized open intent detection to distinguish n-class known intents and a one-class open intent, adhering to the methodologies and results published by TEXTOIR, using identical seeds to match intents in evaluating our DETER framework. OpenMax employs softmax loss and fits a Weibull distribution to classifier logits \cite{bendale2015open}, while MSP predicts known classes based on maximum softmax probabilities, excluding those with probabilities below 0.5 \cite{hendrycks2018baseline}. LOF utilizes density to pinpoint low-density outliers \cite{Breunig2000}, while DOC establishes multiple probability thresholds for known classes through Gaussian fitting \cite{shu-etal-2017-doc}. DeepUnk integrates margin loss with LOF for adept identification of unknown classes \cite{lin-xu-2019-deep}, and SEG utilizes a large margin Gaussian mixture loss for feature representation, offering a unique approach within deep learning techniques \cite{yan-etal-2020-unknown}.

Diverging in methodology, ADB, a variant of DA-ADB, favours softmax loss over traditional distance-aware concepts ~\cite{DA-ADB_2023, Zhang_2021}. Expanding on intent representations, the (K+1)-way method merges features from two in-domain intents ~\cite{Zhan_2021}. MDF stands out on the outlier detection front by employing a one-class SVM post-Mahalanobis distance feature evaluation ~\cite{xu-etal-2021-unsupervised}. ARPL innovates by maximizing variance between known-class samples and reciprocal point representations, setting its detection threshold at 0.5 ~\cite{Guangyao2022}. KNNCL captures attention by utilizing KNN, enhancing the learning of semantic features crucial for detecting out-of-scope intents ~\cite{zhou-etal-2022-knn}.

%%%%%%%%%%%%%%% 

Training strategies, in particular, often become the defining factor for a model's success. A novel approach that has gained traction involves the infusion of synthetic outliers and the integration of OOS sentences from varied datasets during training \cite{Zhan_2021}. Yet, specific strategies, like threshold introduction, may enhance OOS detection over BiLSTM baseline at the expense of in-domain performance \cite{Hasani_2022}. Among these challenges, deep metric learning emerges as a novel approach in the intent detection space, fostering enhanced intent representation through techniques like triplet networks \cite{Zhang_2023}. Combined with advanced techniques like triplet loss and hard samples, these networks provide new precedents in discriminative intent representation.

\section{Dual Encoder for Threshold-based Re-Classification (DETER)}

This section describes the proposed DETER framework and discusses its internal workings.

Given \textit{K} predefined intent classes $(S_{known})$ in a dialogue system, an intent detection model aims to predict an utterance's class, which may be one of the known intents or an out-of-scope intent. During the training phase, a set of \textit{N-labeled utterances} $(D^{tr}_n)$ is provided for training as shown below in Equations~\ref{eq01} and~\ref{eq02}. This depicts a $K+1$ way classification problem at the test phase. Previous approaches commonly train a K-way classifier for the known intents and then attempt to find the outliers.
\begin{equation}\label{eq01}
  S_{known} = \{C_i\}^K_{i=1}
\end{equation}
\begin{equation}\label{eq02}
  D^{\text{tr}}_n = \{(u_i, C_i)\}^N_{i=1} \mid C_i \in S_{known}
\end{equation}

where $C_i$ is the $i^{th}$ intent class in the in-domain dataset, $K$ is the total number of predefined intent classes, and $u_i$ is the utterance.

\begin{figure*}[!ht]
\begin{center}
  \includegraphics[width=4.3 in]{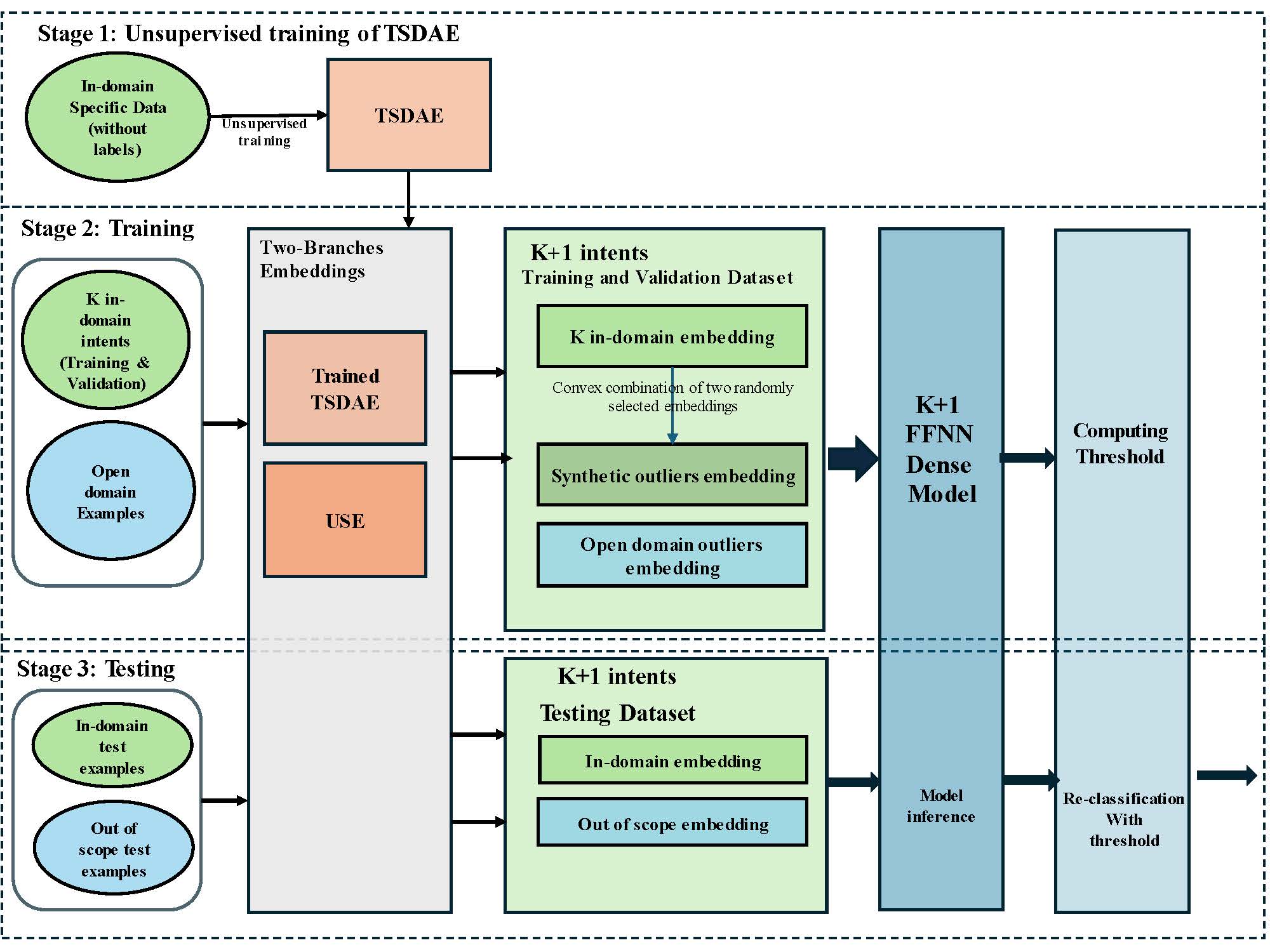}
  \caption{Overview of the proposed Dual Encoder for Threshold-based Re-Classification (DETER).}\label{fig_1}
\end{center}
\end{figure*}

On the other hand, as illustrated in Figure \ref{fig_1}, DETER formalizes a (K + 1)-way classification task in the training phase by constructing out-of-scope examples through self-supervision and open-domain data without making data distribution assumptions. The dual encoder-based trained classifier can be easily used for inference without adaptation or post-processing. In the following subsections, we describe our proposed approach's details, including the dual encoders -- Universal Sentence Encoder (USE) and Transformer-based Denoising AutoEncoder (TSDAE) -- representation learning, construction of pseudo outliers, and training.

\subsection{Universal Sentence Encoder (USE)}
USE is a text embedding framework designed to transform text into high-dimensional vectors. Given a user utterance $u$, the USE embedding function, denoted by \(E_{\text{USE}}\), maps \( u \) to a 512-dimensional vector in \(\mathbb{R}^{512}\), i.e., $E_{\text{USE}}: u \rightarrow \mathbb{R}^{512}$.

These embeddings have proven invaluable for many NLP tasks, including text classification, semantic similarity measurement, and clustering. The pre-trained USE is publicly accessible on Tensorflow-hub and is available in two forms, one trained with a Transformer encoder and the other with a Deep Averaging Network (DAN). The output is a 512-dimensional vector, while the input is the English text of varying length \cite{Daniel_2018}.

\subsection{Transformer-based Denoising AutoEncoder (TSDAE)}
TSDAE provides unsupervised training of the transformer architecture by introducing particular types of noise into the input text and subsequently reconstructing the vector representations (of the noisy input) to their original input values. During the training, TSDAE encodes broken sentences into fixed-sized vectors, and the decoder uses these sentence embeddings to rebuild the original sentence representations. The encoder's sentence embedding must accurately capture the semantics for higher-quality reconstruction.

We tune the TSDAE, built upon a RoBERTa transformer, in an unsupervised mode using the label-free CLINC-150 dataset, achieving domain adaptation \cite{Wang_2021}. At inference, the encoder 
\( E_{\text{TSDAE}}\) transforms variable-length user utterance,\(u\), into a fixed 768-dimensional vector Representation in \(\mathbb{R}^{768}\), $E_{\text{TSDAE}}: u \rightarrow \mathbb{R}^{768}$.

Through noise-induced unsupervised tuning, TSDAE can effectively capture intricate sentence semantics, making it suitable for intent detection tasks and discerning user purposes from variable-length user utterances. 

\subsection{Representation Learning}
To capture rich linguistic and contextual feature representation from user utterances, We employed the above two sentence embedding techniques: TSDAE and USE. Each user utterance, \( u\), is independently processed through both the encoders, yielding a d-dimensional output vector of the TSDAE and USE, as shown in Equation \ref{eq1}.
%
%\begin{equation}\label{eq1}
%  h = E_{\text{TSDAE}}(u) \oplus E_{\text{USE}}(u) \in R^d
%\end{equation}
% + operator is confusing when we use it for concat
%Reformated equation - WR
\begin{equation}\label{eq1}
h(u) = \begin{bmatrix} E_{\text{TSDAE}}(u) \\ E_{\text{USE}}(u) \end{bmatrix}
\end{equation}

The dimensionality \(d\) of \(h(u)\) amounts to 1280, which is attained by concatenating the embedding outputs from TSDAE (768 dimensions) and USE (512 dimensions). As a result, the user utterance transforms into a vector \(h(u)\) residing in \(\mathbb{R}^{1280}\).

Post this representation learning, the training set is mapped to $D^{tr}_n$ in the feature space as shown in equation \ref{eq2}.
%
%\begin{align} \label{eq2}
 %   D^{tr}_n & = \{(h_i, c_i) | h_i \nonumber \\
 %   & = TSDAE(u_i) + USE(u_i), (u_i, c_i) \in D_n\}^N_{i=1}
%\end{align}

% reformated equation- WR
\begin{equation}
\begin{aligned}
\label{eq2}
     D^{tr}_n =  \left\{(h_i, C_i) \mid h_i 
     = \begin{bmatrix} 
            E_{\text{TSDAE}}(u_i)\\
            E_{\text{USE}}(u_i) 
            \end{bmatrix}, \,
        (u_i, c_i) \in D_n \right\}^N_{i=1}
\end{aligned}
\end{equation}

\subsection{Construction of Outliers}

We implement the outlier construction method proposed by \citet{Zhan_2021}, which involves generating two training outliers: synthetic outliers via self-supervision and readily available open-domain outliers.

\paragraph{Synthetic Outliers}
We introduce hard outliers in the feature domain to enhance the model's generalization capability for out-of-scope (unknown) intent classification. These outliers are hypothesized to have representations analogous to established classes. Leveraging this hypothesis, we propose a self-supervised methodology to generate these hard outliers using the training set $D^{tr}_n$ as delineated by \citet{Zhan_2021}.

Specifically, we construct synthetic outliers within the latent space by utilizing convex combinations of inlier features derived from disparate intent classes, as depicted in Equation \ref{eqs1}.
\begin{equation}\label{eqs1}
  h^{oos} = \theta \times h_\beta + (1 - \theta) \times h_\alpha
\end{equation}

Where $h_\alpha$ and $h_\beta$ are dual sentence encoder embeddings of two in-domain utterances randomly sampled from different intent classes in $D^{tr}_n$, and $h^{oos}$ is the constructed synthetic outlier. $\theta$ is chosen arbitrarily from a uniform distribution $U(0; 1)$. To ensure diversity in synthetic Out-Of-Scope (OOS) training examples, we randomly sample $\theta$ within a specified range. This approach allows us to incorporate "hard," "medium," and "easy" examples for each iteration, thereby generating synthetic outliers that contribute to model training.

%%%% equations %%%%

\paragraph{Open-Domain Outliers}
User input in practical dialogue systems can be arbitrary free-form sentences. Hence, learning architectures can easily use the available open-domain outliers to mimic real-world outliers and provide learning signals representing them during training. We thus model arbitrary outliers using the SQuaD 2.0 question-answering dataset \cite{Rajpurkar_2018} \cite{Zhan_2021}.
Combining the above outliers for training provides robustness to our framework for OOS detection. Later, we will study the effect of these outliers on DETER's performance.

%Moreover, we use 211 OOS examples, selected from the open-domain outliers, for development.

\subsection{DETER Training Architecture}
%\sout{In the DETER framework, user utterance embeddings from TSDAE and USE are integrated into a dual-branch deep learning structure for classification. This model refines the static embeddings from USE and TSDAE, ensuring closely aligned latent representations for similar intents. Each branch, dedicated to USE and TSDAE embeddings, incorporates dense layers with ReLU activations \cite{Maas_2013}, leading to a softmax multi-class classification layer.}
%WR

\begin{figure}[!ht]
\begin{center}
{\includegraphics[width=0.45\textwidth]{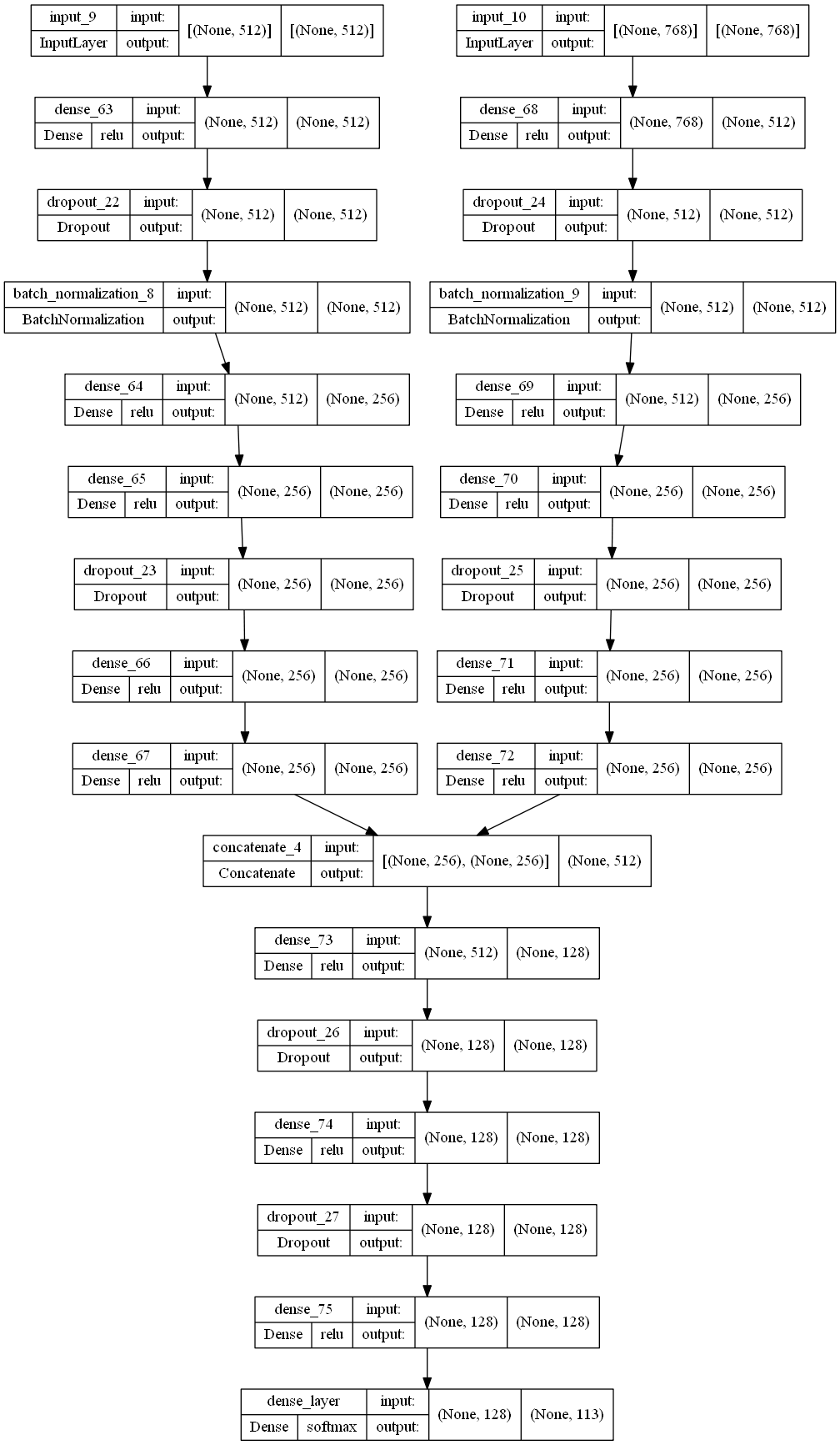}}%
\caption{The model's architecture} \label{fig_3}
\end{center}
\end{figure}

The model architecture comprises two branches: one processing the Universal Sentence Encoder (USE) embedding of the input, while the other handles the Transformer-based TSDAE embedding. This refines the embeddings, aligning latent representations for similar intents. Each branch independently undergoes several dense and dropout layers to diminish the embedding dimension size. Subsequently, the condensed embeddings from both branches are concatenated before being forwarded to a classification head. Figure \ref{fig_3} illustrates the model's architecture. The entire system is optimized using cross-entropy loss against ground-truth intent labels from training data.

The aggregate count of trainable parameters in the proposed models amounts to 1,559,808 (1.5 M), in contrast to 124,645,632 (125 M) as observed in the K+1 method by \citet{Zhan_2021}.
%total number of trainable parameters of \cite{Zhan_2021}. This provides significant optimization in terms of computational efficiency and usability.

\subsection{Re-Classification Threshold}
A key feature of our DETER framework involves enhancing model predictions through potential re-classification to enhance performance. In pursuit of this objective, we adopt the threshold-based approach proposed by \citet{Larson_2019} for the softmax output of the classification model, as illustrated in Equation \ref{eq222}.
\begin{equation}\label{eq222}
  p(c|u) = softmax(Wh + b) \in R^{K+1}
\end{equation}
where $W \in R^{(K+1)\times d}$ and $b \in R^{K+1}$ are the classifier's model parameters.

At inference time, we first take the prediction output of the model, i.e., the class $C_j$ with the most significant value of $p(c = C_j|u)$. However, if the prediction confidence score of the model for the intent associated with an utterance ($I(u)$) is greater than a pre-computed threshold $T$, only then the model output is considered. Else, DETER tags the intent OOS. This is computed as in Equation \ref{eq333}.
\begin{equation}\label{eq333}
I(u) =
\left\{
  \begin{array}{ c l }
    C_j & \quad \textrm{if }  p(c = C_j|u) \geq T \\
    OOS & \quad \textrm{otherwise}
  \end{array}
\right.
\end{equation}

It is noteworthy that the threshold $T \in [0,1]$ is calibrated on the validation set without utilizing any OOS examples. Also, the original OOS examples from the training dataset were not employed during the training and validation phases of the DETER framework. Instead, unselected intents and synthetic outliers were utilized as OOS examples. Furthermore, the threshold was fine-tuned using the provided validation set.

\section{Experimental Setup}
We now present DETER's experimental findings for intent classification. % in-domain (known) and out-of-scope (unknown) intents. 
The main objective is to assign the proper intent labels to utterances in the test set consisting of in-domain (known) and out-of-scope (unknown) intents. We separately report the approaches' macro {\em F1-score} performance for the known and unknown intents. We compare the performance of the proposed DETER framework with the state-of-the-art methodologies in the TEXTOIR framework ~\cite{zhang-etal-2021-textoir}.

\subsection{Dataset}
We evaluated the competing approaches on three real-world datasets, CLINC-150, Banking77, and Stackoverflow, as shown in the table \ref{Table:datasets_details}.

CLINC-150 consists of 150 intent classes across ten domains designed for out-of-scope intent detection. The 1,200 OOS sample texts are split as 100 train, 100 validation, and 1,000 test \cite{Larson_2019}. 
Banking77 is banking-specific data with 77 intents from 13,083 customer queries \cite{Casanueva_2020}.
Stackoverflow features 20 classes, each with 1,000 examples \cite{Xu_2015}.

In this study, we consolidate all 1,200 OOS texts from CLINC-150 into the test set, as our framework does not utilize labelled OOS examples during training \cite{Larson_2019}. Experiments conducted with 25\%, 50\%, and 75\% of dataset intents were repeated five times each to ascertain model robustness and maintain consistent intents for each run.

\subsection{Model Hyper-parameters}
In our experimental setup, we set the batch size, number of epochs, and patience for stopping criteria on validation accuracy as 200, 1000, and 100, respectively. 
We maintain consistency with the epoch count and patience interval as the (K+1)-way method \cite{Zhan_2021} and replicate the experimental setup of the TEXTOIR framework \cite{zhang-etal-2021-textoir}. Specifically, we conduct ten experiments for each dataset, randomly selecting 25\%, 50\%, and 75\% of intent classes from the training set as known classes for training, with the remainder reserved as unknown classes for testing. We utilize 1200 Out-Of-Scope (OOS) examples from the CLINC-150 dataset as OOS test samples for all datasets.

\begin{table*}[!ht]
\begin{center}
    \small
    \begin{tabular}{c|c|cc|cc|cc}
        \toprule 
        \multirow{2}{*}{{Dataset}} & \multirow{2}{*}{{Intent}} & \multicolumn{2}{c}{{Training}}
       & \multicolumn{2}{c}{{Validation}} & \multicolumn{2}{c}{{Testing}}  \\ 
        \cmidrule{3-8} 
        & & {Total} & {25\%} & {Total} & {25\%} & {Total} & {25\%} \\ 
        \midrule
        \midrule 
        \multirow{2}{*}{Clinc-150} & Known & 15,000 & 3,800 & 3,000 & 760 & 4,500 & 1,140\\

        & Unknown & 0 & 11,200 & 0 & 221 & 1,200 & 3,360 + 1,200 \\
        
        \multirow{2}{*}{Banking77} & Known & 9,003 & 2,119 & 1,000 & 234 & 3,080 & 760 \\
        
        & Unknown & 0 & 6,884 & 0 & 221 & 1,200 & 2,320 + 1,200 \\

        \multirow{2}{*}{Stackoverflow} & Known & 12,000 & 3,000 & 2,000 & 500 & 6,000 & 1,500 \\

        & Unknown & 0 & 9,000 & 0 & 221 & 1,200 & 4,500 + 1,200 \\
        
        \bottomrule
    \end{tabular}
     \caption{Total number of examples and the computed examples of 25\% intent ratios.}
\label{Table:datasets_details}
\end{center}
\end{table*}

%Hossam, written down with the updated ranges
%Moreover, the number of randomly selected examples from the open domain is in the range of [50, 8000], and synthetic outliers are in the range of [50, 32000].

%%WR
%\textcolor{red}{
%This range is selected to study the effect of varying amounts of synthetics and open-domain outliers. We empirically observed that the best results were around 500 open-domain and 500 synthetic outliers (as reported). 
%Further, we would like to point out that the (K+1)-way method uses the entire SQuaD 2.0 question-answering dataset (with nearly 100,000 data points) to construct a set of open-domain outliers \cite{Zhan_2021}. Hence, the use of such large numbers of added outliers is well-established in the literature.
%The SQuaD dataset is an open-domain dataset created predominantly from Wikipedia, while the CLINC-150, Stackoverflow, and Banking77 are specific closed-domain datasets. Hence, we posit that the probability/degree of overlap is quite low. Further, please note that the SQuaD dataset has been used previously in the literature \cite{Zhan_2021} as a proxy OOS example for intent classification.}

This range is chosen to analyze the impact of varying amounts of synthetic and open-domain outliers, with optimal results observed around 500 open-domain and 500 synthetic outliers. The (K+1)-way method employs the entire SQuaD 2.0 dataset, comprising nearly 100,000 data points, to generate open-domain outliers \cite{Zhan_2021}, establishing the precedent for incorporating large numbers of outliers. While the SQuaD dataset is open-domain, the CLINC-150, Stackoverflow, and Banking77 datasets are specific closed-domain datasets, minimizing the likelihood of overlap. Additionally, the SQuaD dataset has been previously used as a proxy OOS example for intent classification \cite{Zhan_2021}.

For the model optimization, we use $AdamW$ \cite{Adam} and {\em categorical cross entropy} as a loss function for the multi-class classification model.

We studied the impact of different outliers during model training using varied open-domain and synthetic outliers. Notably, on the validation set, we observed that excessive synthetic outliers adversely affect the open domain performance. Hence, the number of open-domain outliers varied from [50, 4000] and synthetic outliers from [50, 16000]. Further, for re-classification of the model predictions, the threshold $T$, based on the validation set, was set to $0.7$.

\section{Evaluation Results}
We extensively compare our proposed DETER framework with existing OOS intent detection techniques. We consider $3$ different case studies as:

% Wael Edited
\subsection*{Evaluating DETER Against State-of-the-Art} 

We compare our DETER framework with TEXTOIR state-of-the-art open intent detection techniques, namely {\tt (i)} OpenMax ~\cite{bendale2015open}, {\tt (ii)} DOC ~\cite{shu-etal-2017-doc}, {\tt (iii)}ARPL ~\cite{Guangyao2022}, {\tt (iv)} DeepUnk ~\cite{lin-xu-2019-deep}, {\tt (v)} SEG ~\cite{yan-etal-2020-unknown}, {\tt (vi)} (K+1)-way ~\cite{Zhan_2021}, {\tt (vii)} ADB ~\cite{Zhang_2021}, {\tt (viii)} MSP ~\cite{hendrycks2018baseline}, {\tt (ix)} LOF ~\cite{Breunig2000}, {\tt (x)} KNNCL ~\cite{zhou-etal-2022-knn}, {\tt (xii)} MDF ~\cite{xu-etal-2021-unsupervised}, and {\tt (xiii)} DA-ADB ~\cite{DA-ADB_2023}.

For a fair comparison, each technique uses the same BERT model as backbone \cite{zhang-etal-2021-textoir}.
% Wael moved this
Our study uses the TEXTOIR platform to standardize intent selection for training and testing \cite{zhang-etal-2021-textoir}. We used the same seeds as TEXTOIR for consistent results on benchmark datasets to ensure identical intents in DETER.

% wael added
\paragraph{\bf Results:} The results from our proposed approach, evaluated at various known intent ratios (25\%, 50\%, and 75\%) of the total CLINC-150 known intents, are presented in Table \ref{Table:Clinc25-50-75}. Across all datasets and known intent ratios, the DETER framework consistently outperforms other methods in known and unknown intent detection. Notably, in the Clinc-150 and Banking-77 datasets, DETER's scores for unknown intents are particularly commendable. Some techniques like KNNCL and LOF face significant challenges with unknown intents in the Stackoverflow dataset.
%% Sourav Added
%% Wael edited
Notably, many current methods find it challenging to classify known or unknown samples with different training intents accurately. For instance, OSS detection is more efficient in 25\%  and 50\% scenarios, while in-domain classification is superior for 75\%. However, we outperform others in detecting in-domain and OOS in all scenarios.

%In most cases, DETER achieves an impressive F1 score of more than $90\%$ for both in-domain and OOS classification, even with varying training data. 
%This suggests that DETER is especially beneficial in data-scarce situations and adept at handling unfamiliar user inputs.
%Overall, DETER's consistent and superior performance underscores its robustness and capability.

%We observed that incorporating a higher volume of synthetic data enhances model performance under limited training data (as in the 25\% scenario). This is attributed to the model's increased efficiency in delineating the OOS class boundaries when relying on a restricted in-domain dataset. Conversely, with more extensive training data (i.e., 75\% of intents), a balanced dataset proves more beneficial.

% WR
DETER consistently achieves an impressive F1 score exceeding 90\% for in-domain and out-of-scope (OOS) classification, even with varying training data, demonstrating its efficacy in data-scarce scenarios and adaptability to unfamiliar user inputs. The robust and superior performance of DETER underscores its capability. We observed that augmenting synthetic data volume enhances model performance, particularly under limited training data conditions (e.g., 25\% scenario), attributed to improved delineation of OOS class boundaries with restricted in-domain datasets. Conversely, a balanced dataset proves more beneficial with extensive training data (e.g., 75\% of intents).
%%
% Wael edited

% 
%
% Wael edited the table format

\begin{table*}[!ht]
\begin{center}
    \small
    \begin{tabular}{c|c|cc|cc|cc}
        \toprule 
        \multirow{3}{*}{{Dataset}} &\multirow{3}{*}{{Model}} &
         \multicolumn{6}{c}{{Intent Ratio}}\\
          %\cmidrule{3-8}
       & & \multicolumn{2}{c}{{25\%}} & \multicolumn{2}{c}{{50\%}} & \multicolumn{2}{c}{{75\%}} \\ 
        \cmidrule{3-8} 
        & & {Known} & {Unknown} & {Known} & {Unknown} & {Known} & {Unknown} \\ 
        \midrule
          \midrule
        
        \multirow{13}{*}{Clinc-150} & (K+1)-way & 74.02 & 90.27 & 81.52 & 84.25 & 86.72 & 79.59 \\ 
       & ADB & 77.85 & 92.36 & 85.12 & 88.6 & 88.97 & 84.85\\ 
        & ARPL & 73.01 & 89.63 & 80.87 & 81.81 & 86.1 & 74.67\\ 
        & DA-ADB & 79.57 & 93.2 & 85.58 & 90.1 & 88.43 & 86\\ 
        & DOC & 75.46 & 90.78 & 83.84 & 87.45 & 87.91 & 83.87\\ 
        & DeepUnk & 76.95 & 91.61 & 83.3 & 87.48 & 86.57 & 82.67\\ 
        & KNNCL & 78.85 & 93.56 & 83.25 & 87.85 & 86.14 & 82.05\\ 
        & LOF & 77.77 & 91.96 & 83.81 & 87.57 & 87.24 & 82.81\\ 
        & MDF & 49.43 & 84.89 & 61.6 & 62.31 & 72.21 & 51.33\\ 
        & MSP & 51.02 & 59.26 & 72.82 & 63.71 & 83.65 & 63.86\\ 
        & OpenMax & 62.65 & 77.51 & 79.83 & 82.15 & 71.14 & 75.18\\ 
        & SEG & 46.67 & 59.22 & 62.57 & 61.34 & 42.72 & 40.74\\ %\cdashline{2-8}
        & \textbf{Our (DETER)} & \textbf{92.19} & \textbf{98.42} & \textbf{92.02} & \textbf{96.26} & \textbf{92.15} & \textbf{92.51}\\ 
        \midrule
        
               \multirow{13}{*}{{Banking-77}} & (K+1)-way & 67.7 & 82.66 & 77.97 & 72.58 & 85.14 & 59.89\\
        & ADB & 70.92 & 85.05 & 81.39 & 79.43 & 86.44 & 67.34\\
        & ARPL & 62.99 & 83.39 & 77.93 & 71.79 & 85.58 & 61.26\\
        & DA-ADB & 73.05 & 86.57 & 82.54 & 81.93 & 85.93 & 69.37\\
        & DOC & 65.16 & 76.64 & 78.38 & 72.66 & 84.14 & 63.51\\
        & DeepUnk & 64.97 & 76.98 & 75.61 & 67.8 & 81.65 & 50.57\\
        & KNNCL & 65.54 & 79.34 & 75.16 & 67.21 & 81.76 & 51.42\\
        & LOF & 62.89 & 72.64 & 76.51 & 66.81 & 84.15 & 54.19\\
        & MDF & 44.8 & 85.7 & 64.27 & 57.72 & 75.47 & 33.43\\
        & MSP & 50.47 & 39.42 & 73.2 & 46.29 & 84.99 & 46.05\\
        & OpenMax & 53.42 & 48.52 & 75.16 & 55.03 & 85.5 & 53.02\\
        & SEG & 51.48 & 51.58 & 63.85 & 43.03 & 70.1 & 37.22\\ %\cdashline{2-8}
        & \textbf{Our (DETER)} & \textbf{87.45} & \textbf{97.86} & \textbf{88.30} & \textbf{95.44} & \textbf{87.90} & \textbf{93.25}\\
         \toprule
        \multirow{13}{*}{{Stackoverflow}} & (K+1)-way & 50.54 & 52.23 & 70.53 & 51.69 & 81.2 & 45.22\\
        & ADB & 77.62 & 90.96 & 85.32 & 87.7 & 86.91 & 74.1\\
        & ARPL & 60.55 & 72.95 & 78.26 & 73.97 & 85.24 & 62.99\\
        & DA-ADB & 80.87 & 92.65 & 86.71 & 88.86 & 87.66 & 74.55\\
        & DOC & 56.3 & 62.5 & 77.37 & 71.18 & 85.64 & 65.32\\
        & DeepUnk & 47.39 & 36.87 & 67.67 & 35.8 & 80.51 & 34.38\\
        & KNNCL & 41.79 & 15.26 & 61.5 & 8.5 & 76.16 & 7.19\\
        & LOF & 40.92 & 7.14 & 61.71 & 5.18 & 76.31 & 5.22\\
        & MDF & 48.13 & 83.03 & 62.6 & 50.19 & 73.96 & 28.52\\
        & MSP & 42.66 & 11.66 & 66.28 & 26.94 & 81.42 & 37.86\\
        & OpenMax & 47.51 & 34.52 & 69.88 & 46.11 & 82.98 & 49.69\\
        & SEG & 40.44 & 4.19 & 60.14 & 4.72 & 74.24 & 6\\ 
        & \textbf{Our (DETER)} & \textbf{88.16} & \textbf{97.35} & \textbf{88.82} & \textbf{94.71} & \textbf{88.16} & \textbf{90.35}\\ 
        \bottomrule
    \end{tabular}
     \caption{Performance comparison (macro F1-score) of DETER framework for in-domain (known) and OOS (unknown) samples with varying amounts of known training intent ratios (25\%, 50\%, and 75\%) for CLINC-150, Banking-77, and Stackoverflow datasets.}   
%     \caption{Evaluation of DETER framework with different known intent ratios (25\%, 50\%, and 75\%) for CLINC-150, Banking-77, and Stackoverflow datasets. The Macro F1-score for known and unknown intents showcases the model's effectiveness with varying subsets of labelled training data. Bold text indicates the best performance for each dataset.}
      \label{Table:Clinc25-50-75}
\end{center}
\end{table*}

\begin{table*}[!ht]
\begin{center}
    \small
    \begin{tabular}{c|cc|cc|cc}
        \toprule 
        \multirow{3}{*}{{Embedding}} & \multicolumn{6}{c}{{Intent Ratio}}\\
          %\cmidrule{3-8}
       & \multicolumn{2}{c}{{25\%}} & \multicolumn{2}{c}{{50\%}} & \multicolumn{2}{c}{{75\%}} \\ 
        \cmidrule{2-7} 
        & {Known} & {Unknown} & {Known} & {Unknown} & {Known} & {Unknown} \\ 
        \midrule
          \midrule\textbf{TSDAE (Clinc-150) \& USE} & \textbf{87.45} & \textbf{97.86} & \textbf{88.30} & \textbf{95.44} & \textbf{87.90} & \textbf{93.25}\\

        TSDAE(Roberta) \& USE & 65.61 & 89.84 & 70.77 & 78.76 & 84.47 & 83.83 \\ 
 
        TSDAE (Askubuntu) \& USE & 62.43 & 88.52 & 76.23 & 84.61 & 81.77 & 82.09\\ 
        
        TSDAE (Scidocs) \& USE & 56.32 & 88.39 & 74.77 & 84.53 & 82.17 & 81.45\\ 
        
        USE\_only & 52.74 & 90.00 & 76.55 & 85.65 & 81.09 & 77.98\\         
        \bottomrule
    \end{tabular}
     \caption{Performance comparison (macro F1-score) of DETER framework for in-domain (known) and OOS (unknown) samples with varying amounts of known training intent ratios (25\%, 50\%, and 75\%) for Banking-77 dataset using different embedding settings.}
\label{Table:banking25-50-75}
\end{center}
\end{table*}

\subsection*{Assessing DETER with Various Embeddings}
Our investigation revealed that both the standalone USE model and unsupervised training of TSDAE using diverse datasets (Clinc-150, Askubuntu, Scidocs, and original Roberta) exhibited notably inferior performance compared to the proposed joint architecture of DETER.

Ablation experiments conducted on several dual encoder configurations (including USE only, TSDAE only without pre-training using Roberta, TSDAE with unsupervised learning on Clinc-150 combined with USE (DETER), TSDAE with unsupervised learning on Askubuntu combined with USE, and TSDAE with unsupervised learning on Scidocs combined with USE) using the Banking77 dataset are detailed in Table \ref{Table:banking25-50-75}. From the results presented in Table \ref{Table:banking25-50-75}, it is evident that the DETER framework (TSDAE with unsupervised learning on Clinc-150 combined with USE) consistently outperforms other configurations in both known and unknown intent detection tasks. Across all scenarios, DETER achieves an impressive F1 score exceeding 93\% for unknown intents, representing up to a 10\% improvement over the nearest embeddings. Similarly, for known intents, DETER achieves an F1 score surpassing 87\%, exhibiting up to a 22\% improvement over the nearest embeddings.

\subsection*{Model v/s Threshold Performance}
%Graphs for the best results: one graph known and one unknown

% Wael edited=== I moved this to the previous section and changed the section title
%In our study, we utilized the TEXTOIR platform to standardize intent selection for training and testing \cite{zhang-etal-2021-textoir}. We used the same seeds as TEXTOIR for consistent findings on benchmark datasets to ensure identical intents in our DETER framework.

\begin{figure*}[!ht]
\begin{center}
{\includegraphics[width=0.75\textwidth]{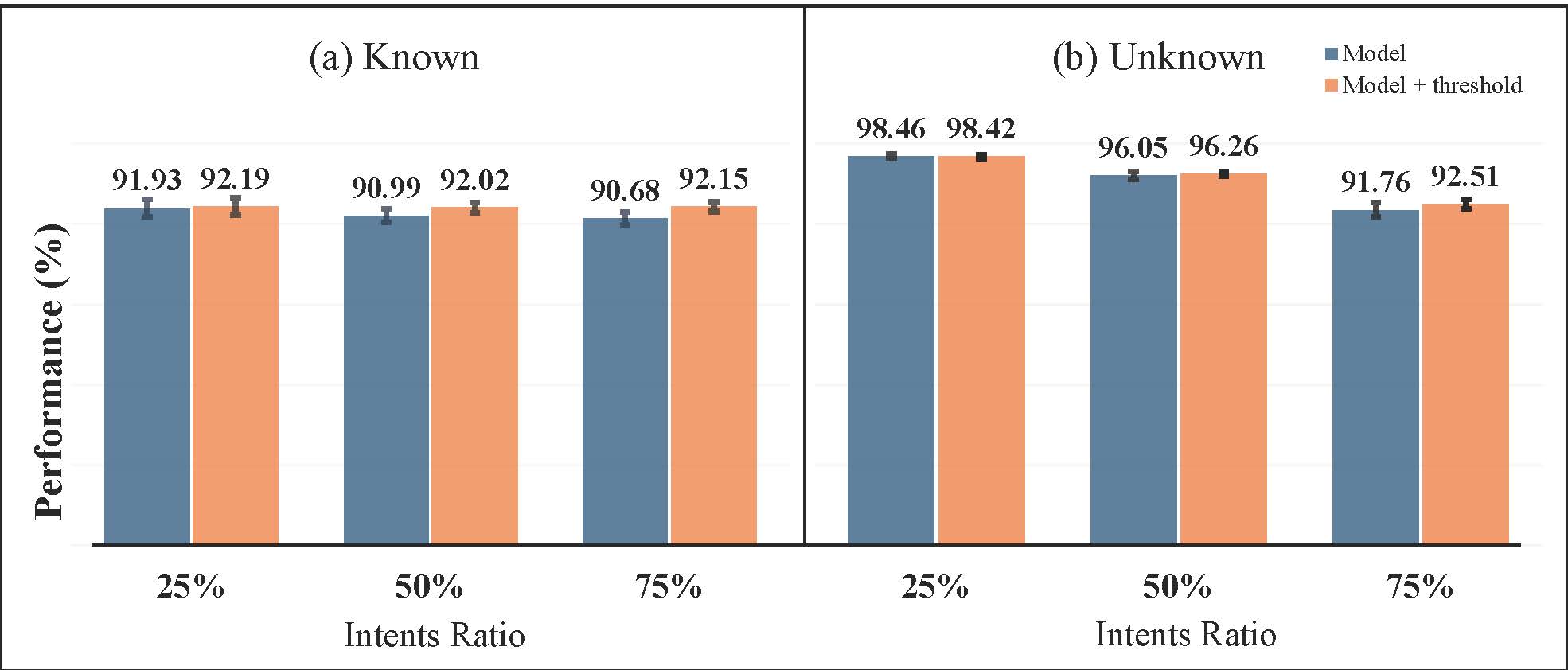}}%
\caption{Performance comparison of the ''Model only'' versus the ''Model with threshold (DETER)'' on CLINC-150 dataset for both (a) known and (b) unknown intents across varying intent ratios
%(25\%, 50\%, and 75\%)
. The error bars display the standard deviation across ten runs.} \label{fig_2}
\end{center}
\end{figure*}

% Wael added 
Figure \ref{fig_2} illustrates a comparative analysis of the F1 performance of the model only (i.e., DETER without the re-classification threshold) and the complete DETER architecture on the CLINC-150 dataset for classifying known and unknown intents. Across all intent proportions, the model with threshold consistently outpaces the standalone version. This emphasizes the positive impact of the DETER framework on model performance. Notably, the model's aptitude in classifying unknown intents is commendable. At the 25\% intent proportion, it reaches a striking 98.46\% F1, demonstrating its proficiency in identifying out-of-scope utterances. Compared to using the model only, the DETER framework displays improved performance and presents lower standard deviations, indicative of model robustness. Although the performance gap narrows with increasing data, the model with the threshold consistently holds an edge. For known intents, the performance remains relatively stable across varying proportions, hovering around 91-92\%, indicating model consistency in recognizing the intents it is trained on, irrespective of data volume. Similar behaviour was demonstrated across the other two datasets, Banking77 and Stackoverflow, underscoring the robustness and versatility of DETER.

\subsection*{Class Weights}
To tackle the class imbalance issue (between in-domain and OOS training data), we evaluated DETER {\em with and without} class weights in this case study. We observed no significant performance difference in the two settings. For example, in the 25\% proportion of CLINC-150, the best F1 results obtained with class weights are 82.07\% for the known and 94.32\% for the unknown. Furthermore, without class weights, the performances are 82.10\% for known and 94.20\% for unknown classes.

\section{Discussion}
% WR added
%Our research introduces the DETER framework that aims at effectively detecting out-of-scope user utterances in dialogue systems. This framework combines dual text encodes (USE and TSDAE), branched deep learning neural network architecture, synthetic and open-domain outliers, and a threshold-based re-classification mechanism.

%Our primary focus has been formulating an efficient and straightforward framework for intent classification. The framework harnesses the combined power of two sentence representations, Universal Sentence Encoder (USE) and Transformer-based Sequential Denoising Auto-Encoder (TSDAE), along with an integration of dense layers.

%WR
Our research presents DETER, a framework for robust out-of-scope user utterance detection. DETER offers a comprehensive approach by combining dual text encoders (USE and TSDAE), a branched deep learning architecture, synthetic and open-domain outliers, and a threshold-based re-classification mechanism. We focus on formulating an efficient intent classification framework, leveraging the combined strength of USE and TSDAE sentence representations integrated with dense layers.

 % Wael Edited this section
\subsection*{DETER's Performance} 

%The efficacy of the DETER framework is reflected in its substantial performance on unknown and known intent classes, as depicted in Figure \ref{fig_2}. 

%\sout{In a practical dialogue system, the frequent confrontation with more outliers than inliers poses a formidable challenge. Factors such as varied user demographic data, users' unfamiliarity with the software, and the constraints of intent classes recognized by the classification model further compound this complexity.} 
DETER's substantial improvements with limited labelled training sets (25\% and 50\% cases in Table \ref{Table:Clinc25-50-75}) indicate its enormous potential in real-life applications. This is particularly significant in an ever-changing landscape of user interactions where adaptive systems are prominent.

Table \ref{Table:Clinc25-50-75} shows that the DETER framework performs best on three benchmark datasets (CLINC-150, Banking77, and Stackoverflow) and outperforms the leading models. In the CLINC-150 dataset, DETER outperforms the state-of-the-art models by 13\%, 6\%, and 3\% for different known percentages (25\%, 50\%, and 75\%) and 5\%, 6\%, and 6\% for different unknown percentages (25\%, 50\%, and 75\%). In the Banking77 dataset, DETER outperforms the state-of-the-art models by 16\%, 7\%, 2\% for different known percentages and 11\%, 13\%, and 24\% for different unknown percentages. Similarly, in the Stackoverflow dataset, DETER outperforms the state-of-the-art models by 7\%, 2\%, 0.5\% for different known percentages and 6\%, 5\%, and 15\% for different unknown percentages.

This performance improvement of DETER is attributed to its unique combination of Universal Sentence Encoder (USE) and Transformer-based Sequential Denoising Auto-Encoder (TSDAE), which offers nuanced sentence embeddings that adeptly bridge universal linguistic traits with domain-specific nuances. Its branched architecture bolsters feature extraction, clearly differentiating intent and OOS utterances. The framework also includes complementary components like synthetic outliers and threshold-based re-classification mechanisms, further enhancing its efficacy.

\subsection*{Synthetic and Open-Domain Outliers} 
Our findings show that synthetic outliers with higher values have a more noticeable impact than open-domain outliers, which extend beyond mere observation. This insight opens up new pathways for research to leverage synthetic data manipulation to enhance model robustness and performance.

% Wael Edited this section
\subsection*{Efficiency, Scalability, and Applicability} 
One of the most salient aspects of this model is its remarkable computational efficiency. The DETER model boasts remarkable computational efficiency with only $1.5 M$ trainable parameters, starkly contrasting the $125 M$ in the model by \citet{Zhan_2021}. This streamlined design, which does not sacrifice performance, enhances scalability and flexibility for deployment across various platforms, even those with limited resources. 
%As DETER can potentially handle larger datasets and intents, it is vital to continually assess its performance and resource metrics to tackle any issues as it scales preemptively.

\subsection*{Performance and Robustness} 
Despite its simplicity of architecture, the model maintained robust performance, substantiating the efficacy of our approach. This performance parity with more complex models underlines our approach's potential to contribute significantly to ongoing research in deep learning and NLP.

% Wael Edited this section
\subsection*{Limitations and Future Work}
Our assessment of the DETER framework encompassed $3$ datasets, providing a broader evaluation spectrum. However, even with this multi-dataset approach, inherent data limitations persist. This  %Such constraints underscore the framework's potential and 
emphasize the necessity of further exploration across even more diversified datasets, including multilingual datasets or languages other than English, to fully evaluate the framework's potential. %\sout{Investigative efforts are needed to determine if the dual encoder setup requires language-specific fine-tuning or additional preprocessing steps. As we envisage advancements in task-oriented conversational systems, we recognize that enhancing the framework may involve iterative refinements, diversifying encoding techniques, and architectural fine-tuning.}
Among future research avenues, few-shot learning is an intriguing domain poised to augment DETER's adaptability and efficiency.

\section{Conclusions}
% WR edit
This work introduces the Dual Encoder for Threshold-based Re-Classification (DETER), designed to detect out-of-scope user utterances in dialogue systems and by employing dual text encoders (USE and TSDAE), branched deep learning neural network architecture, synthetic and open-domain outliers, and a threshold-based re-classification mechanism, the DETER framework was rigorously evaluated using CLINC-150, Banking77, and Stackoverflow datasets. Our findings indicate a promising performance and superiority over the baseline. Through experimentation with synthetic outliers and open-domain data, we found that synthetic outliers exerted a more pronounced influence on the model's efficacy. This insight underscores the significance of adept outlier handling to augment the model's capability. However, introducing class weights had a negligible impact on the results.

While focusing on three benchmark datasets (CLINC-150, Banking77, and Stackoverflow) introduces certain constraints, the overarching conclusions underscore a promising direction for future research. The emphasis lies on optimizing efficiency, scalability, and applicability without sacrificing performance. DETER's evaluation of three benchmark datasets reveals its superiority, outperforming the state-of-the-art models.

This insight suggests a promising avenue for future research. Given that the TSDAE framework utilizes either Roberta or BERT models as its backbone, the possibility of training a multilingual TSDAE using multilingual data arises. This approach, mirroring the original TSDAE framework by Wang et al. (2021), could lead to the development of a multilingual DETER framework capable of handling diverse language datasets.

%%%% Is there any problem with adding the following acknowledgments?
% No acknowledgment in submission version due to anonymity - Sourav

\section{Acknowledgments}
This project has received funding from Enterprise Ireland and the European Union's Horizon 2020 Research and Innovation Programme under the Marie Skłodowska-Curie grant agreement No 847402.

\nocite{*}
\section{Bibliographical References}\label{sec:reference}
\bibliographystyle{lrec-coling2024-natbib}
\bibliography{lrec-coling2024-example}
\end{document}